\documentclass{article}
\usepackage[preprint]{log_2026}

\makeatletter
\renewcommand{\@adddotafter}[1]{#1}
\makeatother

\usepackage{booktabs}
\usepackage{graphicx}
\usepackage{amssymb}     
\usepackage{pifont}      
\usepackage{algorithm}
\usepackage{tabularx}
\usepackage{multirow}
\usepackage[numbers,compress,sort]{natbib}
\usepackage{newtxtext}

\graphicspath{{figs/}}

\newcommand{\edgebank}{\textsc{EdgeBank}}
\newcommand{\slade}{\textsc{SLADE}}
\newcommand{\midas}{\textsc{MIDAS-F}}
\newcommand{\isoforest}{\textsc{IsoForest}}
\newcommand{\simplebase}{\textsc{SimpleCount}}

\title[Beat the Counter First]{Beat the Counter First: A Baseline for
Temporal-Graph Anomaly Detectors}

\author[Ahmed and Shafi]{%
  Omair Shafi Ahmed \\
  Microsoft \\
  \email{oshafiahmed@microsoft.com}
  \And
  Zohair Shafi \\
  Northeastern University \\
  \email{shafi.z@northeastern.edu}
}

\begin{document}

\maketitle

\begin{abstract}
Progress in streaming, edge-level graph anomaly detection (GAD) has been
marked by increasingly elaborate architectures, from count-min-sketch
chi-square tests to memory-augmented attention networks. Yet the empirical
gains attributable to this added complexity have not been systematically evaluated. We
propose \simplebase{}, a reference with no parameter fitting that selects one
scalar feature per dataset from a fixed pool of counts, recencies,
first-occurrence indicators, and count-derived transforms. We compare
\simplebase{} with two temporal-graph detector models and an \isoforest{}
control fitted to the complete feature vector across five public datasets and
one synthetic dataset.
\simplebase{} \textit{matches or exceeds \slade{} on three of six datasets and
exceeds \isoforest{} on all six}. We report paired statistical tests and
five-seed \slade{} evaluations. \slade{} requires \textit{23 to
133$\times$ more wall-clock time} than \simplebase{}. On Synth-Triangle and
an additional Synth-Quad probe, pre-event structural scores recover the planted
signal at AUC up to $0.955$, while all evaluated detector models remain near
random. The benefit of complexity is dataset-dependent, and every claimed gain
should be reported against a strong one-feature reference together with its
compute cost.

\vspace{0.75em}\noindent\textbf{Keywords:} temporal graphs, dynamic graph
anomaly detection, shortcut learning, simple baselines, compute-efficient
graph learning, benchmark evaluation.
\end{abstract}

\section{Introduction}
\label{sec:intro}

Streaming-graph anomaly detection - the task of flagging suspicious edges
$(u,v,t)$ as they arrive in a temporal interaction graph - has followed a
familiar trajectory in the literature: each method enriches the per-edge
representation, from count-min-sketch chi-square tests
(\midas{}~\cite{bhatia2022midas}) through temporal-graph attention
(TGAT~\cite{xu2020tgat}) and memory networks (TGN~\cite{rossi2020tgn}),
to recent self-supervised detector models such as
\slade{}~\cite{lee2024slade}. While the improvements are real, how much of
each gain comes from the added architectural complexity is unclear.

Elsewhere in graph learning, strong baselines have challenged similar gains.
\edgebank{}~\cite{poursafaei2022edgebank} showed that a simple memorization
rule---score 1 if an edge has been seen before,
0 otherwise---rivals TGN, TGAT, CAWN, and JODIE on temporal link prediction.
An MLP mixer~\cite{cong2023graphmixer} later matched attention-heavy models
on standard temporal link-prediction benchmarks. GADBench~\cite{gadbench2023}
found that tree ensembles with simple neighbourhood aggregation outperform
specialised GNN anomaly detectors on \emph{static} graphs. All three results
concern tasks like link prediction and static node-level anomaly detection,
where recurring edges or structural patterns reward low-complexity features.
Neither settles the case for \emph{temporal edge-level} anomaly detection,
where labels target atypical dynamic behaviour rather than future-edge
existence, and where the temporal dimension might plausibly demand the
full machinery of attention and memory.

\simplebase{} puts that assumption to a direct test. We compare the
count-sketch-based \midas{} and self-supervised temporal-graph attention
network \slade{} against a low-complexity reference that selects one scalar
feature per dataset. We also fit \isoforest{} to the complete 14-feature
vector to test whether a nonlinear combination improves over selecting one
feature. This lets us quantify what added model complexity buys in accuracy
and what it costs in compute. We report a separate two-dataset GraphMixer
check~\citep{cong2023graphmixer} because its downstream classifier uses anomaly
labels.

Our results connect to a growing body of work, including an analysis by
\citet{heeg2025weisfeiler} showing that temporal GNNs can fail to distinguish
temporal structures. Strong performance from simple endpoint and pair features
is consistent with \emph{shortcut learning}~\citep{geirhos2020shortcut}
rather than structural reasoning.

Our central finding is that a reference with no parameter fitting is
competitive with a state-of-the-art temporal-graph detector model at one to
two orders of magnitude less compute. Concretely, we show that:

\begin{itemize}
    \item From a performance perspective, across five public datasets and one
    synthetic dataset, \simplebase{} matches or beats \slade{} on three
    datasets and exceeds \isoforest{} on all six in terms of AUC.

    \item From a computational cost perspective, \simplebase{} uses
    $23$--$133\times$ less wall time than \slade{}, with a mean of
    $72\times$ across all six datasets.

    \item From a simplicity and explainability perspective, \simplebase{}
    detections are directly traceable to an observed count, recency, or
    first-occurrence value, without requiring an explanation method such as
    SHAP~\citep{lundberg2017unified}. Models based on feature ensembles or
    learned latent node states do not provide this one-feature attribution.
    
\end{itemize}

Our central claim is that \textbf{the benefit of complexity is
dataset-dependent, and every claimed gain should be reported against a strong
one-feature reference together with its compute cost.} We distil the audit
into five recommendations for benchmark authors and reviewers in
Appendix~\ref{sec:recommendations}.

\section{Related Work}
\label{sec:related}

This work connects two lines of research: temporal-graph anomaly detection and
benchmark studies showing the value of strong simple baselines. We apply this
baseline-first perspective to temporal edge-level anomaly detection and add
paired inference and compute-cost accounting.

\paragraph{Temporal-graph anomaly detection.}
The streaming-graph anomaly-detection subfield has produced
\midas{}~\cite{bhatia2020midas,bhatia2022midas} (count-min-sketch
chi-square), SedAnSpot~\cite{eswaran2018sedanspot},
FFADE~\cite{chang2021ffade}, and most recently
\slade{}~\cite{lee2024slade}, which combines TGN-style node
memory~\cite{rossi2020tgn} with a GRU updater, a TGAT memory
generator~\cite{xu2020tgat}, and node-wise self-supervised objectives.
It is formulated for dynamic node anomaly detection; its reference
evaluator produces one anomaly score per interaction from the source
node's memory-recovery and memory-drift similarities. The SLADE study
evaluates Wikipedia, Reddit, Bitcoin-Alpha, and Bitcoin-OTC~\cite{kumar2019jodie,kumar2016edge}. 

\paragraph{Simple baselines.}
The closest prior result is \edgebank{}~\cite{poursafaei2022edgebank}, which
showed that a hash-set of seen edges matches or outperforms TGN, TGAT, CAWN,
and JODIE on multiple temporal link-prediction benchmarks. In our setting, the
\edgebank{}-$\infty$ rule enters as the \textsf{is\_new\_pair} feature.
\citet{cong2023graphmixer} extended this critique to broader temporal networks,
showing that an MLP mixer matches attention-based methods on standard temporal
link-prediction benchmarks. Recent studies likewise found that recency and
popularity heuristics rival neural models~\citep{heuristics2025temporal} and
that negative-sampling artifacts can inflate temporal-graph link-prediction
results~\citep{tgmfail2023global}. These works target link prediction. We study
temporal edge-level anomaly detection, where labels identify atypical dynamic
behaviour and compute efficiency directly affects detection latency.

\paragraph{Corrective benchmark critiques in graph learning.}
A parallel line of work has repeatedly found that simple methods rival
specialized models under careful evaluation. GADBench~\citep{gadbench2023}
showed that tree ensembles with simple neighbourhood aggregation outperform
specialized GNN anomaly detectors on static graphs.
\citet{threerevisits2023} exposed benchmark limitations in node-level graph
anomaly detection and found that message passing can reduce performance.
Most relevant here, \citet{heeg2025weisfeiler} showed that temporal GNNs can
fail to distinguish certain temporal structures, complementing our finding that
\slade{} does not recover the planted multi-hop patterns in our synthetic
probes.

\paragraph{Parsimony audits in machine learning.}
\citet{lipton2018troubling} warned that empirical gains attributed to
new methods often stem from hyperparameter tuning rather than the methods
themselves. We apply the same principle by pairing each accuracy gain with the
compute spent to obtain it
(Section~\ref{sec:results}, Figure~\ref{fig:cost-benefit}).

\section{Methodology}
\label{sec:methods}

Our measurement protocol has four components: (1) a low-complexity
one-feature reference, (2) chronological feature selection that prevents
optimistic bias in that reference, (3) paired significance tests, and (4)
synthetic probes that test whether the evaluated detector models identify
known planted graph patterns.

\subsection{One-feature reference pipeline}
\label{sec:methods:selection}

For every incoming edge $(u,v,t)$, a single chronological pass maintains three
types of state. First, it maintains lifetime counts for source $u$,
destination $v$, and the ordered source--destination pair $(u,v)$. Second, it
maintains counts for the same source, destination, and pair within the current
time bucket; these counts reset when the stream enters a new bucket. Third, it
records the most recent timestamp at which each source, destination, and pair
appeared. All features are read before the current edge updates this state, so
the feature vector uses only prior edges.

\begin{algorithm}[t]
\caption{Chronological simple-feature extraction.}
\label{alg:features}
\small
\setlength{\tabcolsep}{0pt}
\renewcommand{\arraystretch}{1.08}
\begin{tabularx}{\linewidth}{@{}>{\raggedleft\arraybackslash}p{2.8em}@{\hspace{0.75em}}X@{}}
\textsc{in} & Time-sorted edges $(u_k,v_k,t_k,b_k)_{k=1}^{n}$ and unseen-recency sentinel $M$. \\
\textsc{state} & For key type $x\in\{p,s,d\}$ (pair, source, destination):
lifetime count $C_x$, current-bucket count $C_x^{(b)}$, and last-seen time
$T_x$. Initialize all counts to $0$, all times to $\bot$, and
$S,m\gets0$, $b_{\mathrm{prev}}\gets\bot$. \\
\addlinespace[2pt]
1 & \textbf{for} $k=1,\ldots,n$ \textbf{do} \\
2 & \quad Set $q_p=(u_k,v_k)$, $q_s=u_k$, and $q_d=v_k$. \\
3 & \quad \textbf{if} $b_k\ne b_{\mathrm{prev}}$: reset every
$C_x^{(b)}$ and set $b_{\mathrm{prev}}\gets b_k$. \\
4 & \quad \textbf{for each} $x\in\{p,s,d\}$:
$n_x\gets[C_x[q_x]=0]$ and $\Delta_x\gets t_k-T_x[q_x]$ if seen,
else $M$. \\
5 & \quad Set $\bar C\gets S/\max(1,m)$. \\
6 & \quad \textbf{emit}
$\begin{aligned}[t]
\phi_k=(&n_p,n_s,n_d,\ C_p[q_p],C_s[q_s],C_d[q_d],\\[-1pt]
        &C_p^{(b)}[q_p],C_s^{(b)}[q_s],C_d^{(b)}[q_d],\ \Delta_p,\Delta_s,\Delta_d,\\[-1pt]
        &\log(1+C_p^{(b)}[q_p]),\ C_p^{(b)}[q_p]/\max(1,\bar C)).
\end{aligned}$ \\
7 & \quad Update $S\gets S+C_p^{(b)}[q_p]$ and $m\gets m+1$. \\
8 & \quad \textbf{for each} $x\in\{p,s,d\}$: increment $C_x[q_x]$
and $C_x^{(b)}[q_x]$; set $T_x[q_x]\gets t_k$. \\
9 & \textbf{end for} \\
\end{tabularx}
\end{algorithm}

For the five real interaction datasets and synthetic dataset, $b_k$
assigns each timestamp to one of $B=1000$ equal-width intervals spanning
the observed timestamp range. In all
cases, every feature for edge $k$ is computed before edge $k$ updates the
maintained state, so the feature vector depends only on earlier edges.

When a source, destination, or pair appears for the first time, its recency is
undefined because no previous timestamp exists. We encode this missing value as
$10^9$ to keep the feature vector finite and numerical. Separate binary
first-occurrence indicators distinguish this case from a previously observed
key with a long time gap. The value $10^9$ is an implementation convention,
not a model parameter. Feature extraction processes the chronologically
ordered edge stream once, computes all features before updating the state, and
therefore runs in $O(|E|)$ time.

For each of the six datasets, we (i) compute the training-slice AUC of every
candidate feature; (ii) select the feature with the highest effective training
AUC, defined as
$\max(\text{AUC}_{\text{train}}, 1-\text{AUC}_{\text{train}})$; and
(iii) report that feature's orientation-invariant test separability as
$\max(\text{AUC}_{\text{test}},1-\text{AUC}_{\text{test}})$. 

\subsection{Paired Inference}
\label{sec:methods:delong}

Within each dataset, the methods being compared assign scores to the same
held-out test edges and are evaluated against the same labels. Their ROC-AUC
estimates are therefore statistically dependent rather than independent. We
account for this dependence with paired DeLong correlated-ROC
tests~\citep{delong1988comparing} and a paired permutation test.
Section~\ref{sec:setup} describes their application.

\subsection{Synth-Triangle Generator}
\label{sec:methods:synth}

Synth-Triangle tests whether a detector model can recover a known structural
pattern. The common-neighbor score verifies that the planted pattern is
detectable; it is not part of the detector-model comparison.

We generate a temporal network with $n=2000$ nodes and $m=80{,}000$ edges.
Edges arrive according to a homogeneous Poisson process, and nodes have equal
expected interaction rates. After a warm-up period, we replace a fraction
$p=0.01$ of the stream with $pm=800$ planted anomalies.

Each anomalous edge $(a,c)$ closes an existing two-hop path $a-b-c$. We verify
that this pattern is recoverable using the pre-event common-neighbor score

\[
s_{\mathrm{CN}}(a,c,t)
=
\left|N_{<t}(a)\cap N_{<t}(c)\right|,
\]

where $N_{<t}(x)$ contains only neighbors observed before time $t$. The score
uses no anomaly labels or future edges. The experiment therefore tests whether
the evaluated detector models recover a planted structure that is known to be
available in the graph history.

\section{Experimental Setup}
\label{sec:setup}

All methods are evaluated on the same edge streams, chronological split, and
evaluation metric. We use ``pp'' to denote percentage points.
Evaluation code and synthetic generators are available at
\url{https://github.com/Omairss/beat_the_counter_first}.

\subsection{Datasets}
We evaluate on five public datasets and one generated probe
(Table~\ref{tab:datasets}). Wikipedia, MOOC, and Reddit come from the
JODIE releases~\cite{kumar2019jodie}; Bitcoin-Alpha and Bitcoin-OTC are
signed trust networks~\cite{kumar2016edge}; and Synth-Triangle is
generated as defined in Section~\ref{sec:methods:synth}.

\subsection{Statistical tests}
We use the paired DeLong test for Wikipedia, Bitcoin-Alpha, Bitcoin-OTC,
MOOC, and Synth-Triangle. Reddit requires separate treatment because the
train-selected direction of the one-feature score reverses on the test slice.
We therefore use a paired permutation test for Reddit because the reported
statistic is the difference in orientation-invariant effective AUC; details are
given in Appendix~\ref{app:reddit-permutation}.
Both tests
compare methods on identical test edges and labels. Their implementation is
described in Appendix~\ref{app:eval}.

\begin{table}[t]
\caption{Datasets. Positive rate is the fraction of labelled anomalous edges
in the full stream. \emph{Selected feature} is chosen on the chronological
training slice (Section~\ref{sec:methods:selection}). All six datasets use
$B=1000$ equal-width intervals. Their sizes range from $24$K to $672$K edges,
and their anomaly rates range from $0.05\%$ to $7.22\%$.}
\label{tab:datasets}
\centering
\small
\begin{tabular}{lrrl}
\toprule
\textbf{Dataset} & $|E|$ & \textbf{Pos rate} & \textbf{Selected feature} \\
\midrule
Wikipedia      & 157\,474 & 0.14\,\% & $c_u$ (inverted) \\
Bitcoin-Alpha  &  24\,186 & 3.61\,\% & $c_v$ \\
Bitcoin-OTC    &  35\,592 & 7.22\,\% & $\Delta t_v$ (inverted) \\
MOOC           & 411\,749 & 0.99\,\% & $c_u$ (inverted) \\
Reddit         & 672\,447 & 0.05\,\% & $c_u$ \\
Synth-Triangle &  80\,000 & 1.00\,\% & $c_v$ \\
\bottomrule
\end{tabular}
\end{table}

The positive rates in our benchmarks span two orders of magnitude
(0.05\%-7.22\%). ROC-AUC is the standard ranking metric for
unsupervised anomaly detection under such imbalance, and our paired
rank-based inference (Section~\ref{sec:methods:delong}) remains valid
regardless of base rate because it operates on per-edge score orderings rather
than threshold-dependent metrics.

\subsection{Compared methods}

We evaluate four methods on every dataset: \simplebase{}, \midas{}, \slade{},
and \isoforest{}. Random guessing provides a chance reference in the
supplementary artefact.

\begin{itemize}

\item \textbf{\simplebase{}}: one scalar feature selected from the fixed
14-feature count, recency, novelty, and count-derived pool on the
chronological training slice (Section~\ref{sec:methods:selection}). The
five real interaction datasets and Synth-Triangle use  $B=1000$ equal-width
intervals over the observed timestamp range.

\item \textbf{\slade{}}: the self-supervised temporal-graph detector proposed
by \citet{lee2024slade}. We use the authors' implementation and each dataset's
recommended configuration. Each run is trained for 10 epochs on one NVIDIA
H100 GPU; reported AUCs are means across seeds 0--4.

\item \textbf{\midas{}}: We use a NumPy port of the C++ MIDAS-F reference
implementation~\citep{bhatia2022midas}. We hold the repository demo
configuration fixed across all six datasets: two hash rows and 1024 columns per
Count--Min Sketch, a filtering threshold of $10^3$, and a temporal decay factor
of $0.5$.

\item \textbf{\isoforest{}}: an unsupervised nonlinear baseline
\citep{liu2008isolation} fitted to the complete 14-dimensional feature
vector on the chronological training slice and evaluated on the test
slice. We use 200 trees with automatic subsampling and contamination
settings. This baseline tests whether jointly modelling all engineered
features improves over selecting one scalar feature.

\end{itemize}

\paragraph{Separate supervised GraphMixer check.}
We additionally evaluate GraphMixer~\citep{cong2023graphmixer} on
Wikipedia and Reddit using the DyGLib dynamic-node-classification
pipeline: temporal-link pretraining followed by a downstream classifier
trained directly on anomaly labels. Because this supervised classifier uses labels to
fit its decision function, it's evaluated seperately. 
In contrast, \midas{}, \isoforest{}, and
\slade{} do not train on anomaly labels, while \simplebase{} uses
training labels only to select one scalar feature from a fixed pool.
GraphMixer therefore receives materially stronger supervision, so
placing its AUCs in the same ranking would not be a like-for-like
comparison. (Appendix~\ref{app:graphmixer}).

\paragraph{\edgebank{} equivalence.}
The \edgebank{}-$\infty$ memorization baseline
\cite{poursafaei2022edgebank} corresponds to our
\textsf{is\_new\_pair} feature up to orientation: \edgebank{} scores
previously seen pairs high for link prediction, whereas
\textsf{is\_new\_pair} scores novel pairs high for anomaly detection.
Its per-dataset AUC appears in the per-feature breakdown
(Figure~\ref{fig:perfeat}).

Additional evaluation details are given in Appendix~\ref{app:eval}.

\section{Results}
\label{sec:results}

\subsection{Per-dataset AUC across all methods}

Table~\ref{tab:master} reports the effective test ROC-AUC of the four methods
on every dataset across our six-benchmark suite.

\begin{table}[t]
\caption{Effective test ROC-AUC under the chronological 85/15 protocol.
Bold indicates the per-dataset best. \slade{} is run at the repository's
recommended configuration for each dataset; its values are five-seed means.}
\label{tab:master}
\centering
\footnotesize
\setlength{\tabcolsep}{4pt}
\begin{tabular}{lrrrr}
\toprule
\textbf{Dataset} & \textbf{\midas{}} & \textbf{\isoforest{}}
& \textbf{\simplebase{}} & \textbf{\slade{}} \\
\midrule
Wikipedia      & 0.535 & 0.589 & \textbf{0.887} & 0.886 \\
Bitcoin-Alpha  & 0.641 & 0.635 & 0.664 & \textbf{0.766} \\
Bitcoin-OTC    & 0.655 & 0.602 & 0.691 & \textbf{0.772} \\
MOOC           & 0.661 & 0.587 & \textbf{0.727} & 0.602 \\
Reddit         & 0.514 & 0.551 & 0.618 & \textbf{0.755} \\
Synth-Triangle & 0.507 & 0.529 & \textbf{0.568} & 0.522 \\
\bottomrule
\end{tabular}
\end{table}

Across these six benchmarks, \simplebase{} matches or exceeds
\slade{} on Wikipedia, MOOC, and Synth-Triangle (3/6 datasets), while
\slade{} wins on Bitcoin-Alpha, Bitcoin-OTC, and Reddit. \simplebase{}
also exceeds \isoforest{} on all six datasets, showing that a nonlinear
combination of the complete 14-feature vector does not improve on the selected
scalar feature in this suite. Neither \midas{} nor \isoforest{} is the
top-performing method on any dataset. The Wikipedia result is effectively a
tie ($0.887$ versus \slade{}'s $0.886$), and every comparison uses \slade{}
at its documented high-performance setting. For the \simplebase{}-versus-\slade{}
inference, five datasets use paired DeLong tests; Reddit uses the paired
permutation test. Wikipedia and Synth-Triangle are not significant;
the four significant comparisons survive a Holm--Bonferroni correction.
\midas{} is not significantly different from \simplebase{} on
Bitcoin-Alpha.
Figure~\ref{fig:delong} reports the paired effect sizes and confidence
intervals for the five datasets evaluated with DeLong.

\subsection{Complexity-benefit: accuracy gain versus compute premium}
\label{sec:results:cost}

\begin{figure}[t]
\centering
\includegraphics[width=0.85\linewidth]{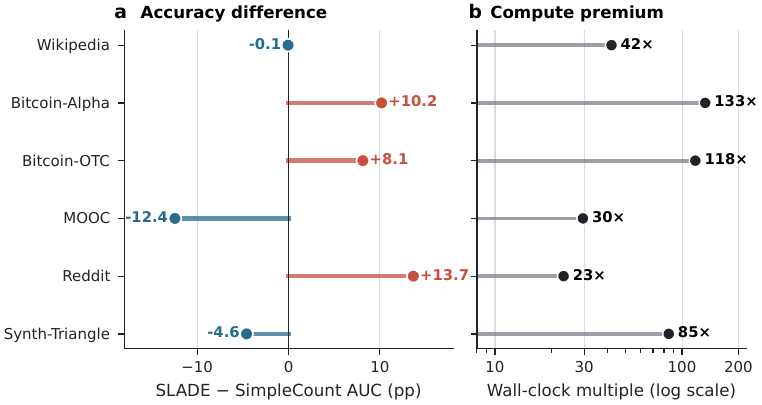}
\caption{Complexity-benefit per dataset. Left: AUC gap (\slade{} $-$
\simplebase{}, percentage points; positive values indicate \slade{} is
ahead). Right: wall-clock cost multiple (\slade{} / \simplebase{},
log scale; mean across six datasets is $72\times$).}
\label{fig:cost-benefit}
\end{figure}

Figure~\ref{fig:cost-benefit} reports the central complexity-benefit finding.
\slade{} uses $23$--$133\times$ the wall time of \simplebase{}, with a mean
of $72\times$ across all six datasets. On the three datasets where \slade{}
leads, the mean cost is $91\times$ for a mean AUC gain of $10.66$\,pp. These
wall-clock measurements use the published implementations.

\subsection{Multi-seed variance}

Across five seeds, \slade{}'s sample standard deviation ranges from
$0.002$ to $0.012$ (Figure~\ref{fig:multiseed}). Winner direction is
stable on five datasets; on near-tied Wikipedia, the numerical ordering
between \slade{} and \simplebase{} changes across seeds.

\subsection{Recoverable local structure that tested methods miss}

\begin{figure}[t]
\centering
\includegraphics[width=0.85\linewidth]{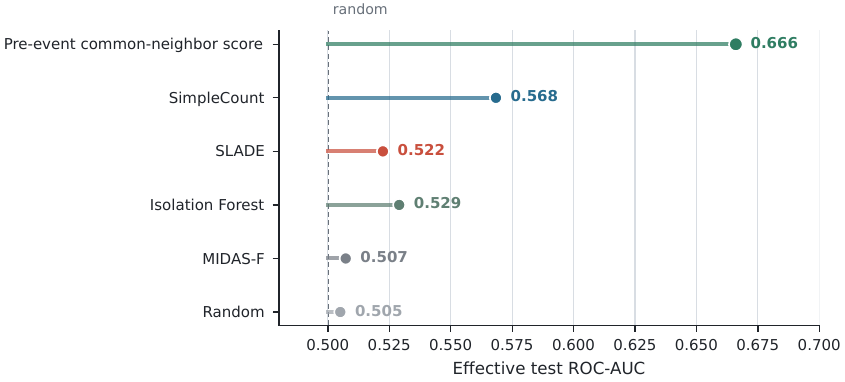}
\caption{Synth-Triangle: feature-based and neural models (Random,
\midas{}, \isoforest{}, \slade{}, \simplebase{}) all operate at or
near chance ($0.50$--$0.57$). A generator-informed pre-event
common-neighbor score attains AUC $= 0.666$ (top row), confirming that
the planted closure structure is recoverable even when \slade{} does not
recover it. The reference line indicates random chance.}
\label{fig:synth}
\end{figure}

On Synth-Triangle, an elementary two-hop neighbourhood score separates
the anomalies, but every method we tested (\midas{},
\simplebase{}, \slade{}, \isoforest{}) operates near chance ($0.50$--$0.57$;
Figure~\ref{fig:synth}). \simplebase{} reaches $0.568$ from a residual
$\deg \ge 1$ generator bias. Because the planting mechanism is known,
we define a pre-event common-neighbor score
\[
\mathrm{score}(u,v,t) = |N(u,t) \cap N(v,t)|,
\]
where $N(x,t)$ contains interactions strictly before time $t$. The
score uses no labels or future edges; it is generator-informed because
its functional form is chosen with knowledge of the planted closure
pattern. It attains AUC $= 0.666$ (mean common-neighbor count $3.66$ on
positives versus $2.64$ on negatives). Thus, the planted two-hop
structure is recoverable from prior edges. The $+14$\,pp gap shows that
\slade{} does not extract this local multi-hop structure.

\paragraph{A second closure pattern: 4-cycles.}
To test whether the generator-informed score result is specific to triangle
closure, we repeat the construction with 4-cycle-closing anomalies
(\textit{Synth-Quad}: 8000 nodes, 40\,000 edges, $2\%$ positives; each
anomaly $a\to d$ closes a sampled length-3 path $a-b-c-d$). Every
tested feature-based and neural model is again near chance
(\simplebase{} $0.557$, \midas{} $0.521$,
\isoforest{} $0.528$, \slade{} $0.506$). Crucially, the 2-hop
common-neighbor score that separates Synth-Triangle is now \emph{also}
near chance ($0.508$): a 4-cycle leaves no 2-hop signature. A
generator-informed pre-event length-3 path-count score,
\[
\mathrm{score}(u,v,t) = \sum_{b \in N(u,t)} |N(b,t) \cap N(v,t)|,
\]
recovers the anomalies at AUC $= 0.955$, a paired-DeLong margin of
$+0.45$ over \slade{} ($z = 23.6$, $p < 10^{-4}$) and $+0.40$ over
\simplebase{} ($z = 20.9$). Like the common-neighbor score, this score
uses only prior edges and no labels; its form is chosen from knowledge
of the generator. Thus, each planted closure pattern is recoverable from prior
edges, while \slade{} recovers neither. The common-neighbor score that
separates triangles does not separate 4-cycles.

\subsection{Per-feature ablation and operational metrics}

The per-feature AUC heatmap (Appendix Figure~\ref{fig:perfeat}) shows
that source count $c_u$ is selected on Wikipedia, MOOC, and Reddit,
while destination count $c_v$ is selected on Bitcoin-Alpha and
Synth-Triangle. Wikipedia and MOOC use the inverted count direction;
Reddit selects source count $c_u$. In implementation-level
throughput measurements, \simplebase{} processes approximately
$18\,000$--$37\,000$\,edges/sec on one CPU core, while \slade{} processes
approximately $140$--$1\,560$\,edges/sec on an H100 GPU.

\section{Discussion}
\label{sec:discussion}

\subsection{When is the complexity premium justified?}

The value of added complexity varies by dataset. On Bitcoin-Alpha and
Bitcoin-OTC, \slade{} gains $8$--$10$\,pp AUC and the graphs contain only
$24$K--$35$K edges. On Wikipedia, MOOC, and Synth-Triangle, the added compute
does not improve AUC. Reddit presents the strongest trade-off: \slade{} gains
$13.7$\,pp at a $23\times$ wall-clock premium. These results show where the
additional compute buys accuracy and where it does not.

\begin{figure}[t]
\centering
\includegraphics[width=0.65\linewidth]{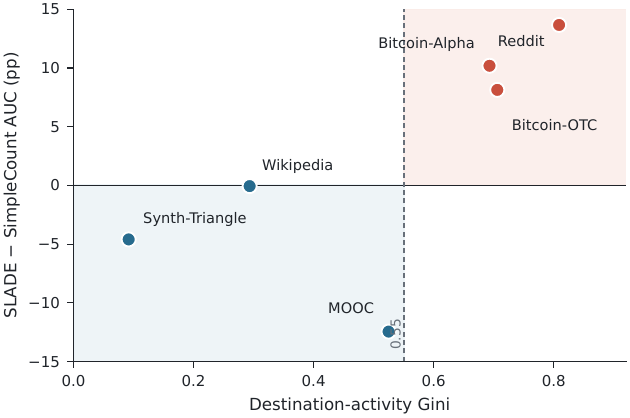}
\caption{AUC gap (\slade{} $-$ \simplebase{}, percentage points) against
destination-activity Gini. Positive values indicate that \slade{} leads;
negative values indicate that \simplebase{} leads.}
\label{fig:regime}
\end{figure}

\paragraph{Dataset structure.}
\label{sec:regime}
Figure~\ref{fig:regime} shows a split at destination-activity Gini $0.55$:
\slade{} leads on the three higher-Gini
datasets, while \simplebase{} matches or exceeds it on the three lower-Gini
datasets. A plausible explanation is that concentrated destination activity
creates hub-related interaction patterns available to attention mechanisms but
absent from a one-dimensional feature. Appendix~\ref{app:regime} gives the
formal definition.

\subsection{Why does MOOC favour \simplebase{}}
The MOOC dropout signal is concentrated in low-activity users. The selected
score is the inverted lifetime source interaction count, $-c_u$, which assigns
higher anomaly scores to interactions from users with fewer prior interactions.
Users who leave after very few sessions are labelled as dropouts, so this
feature captures the label pattern directly. \slade{}'s contrastive objective
encourages node representations to drift minimally over time and may suppress
these activity differences. This interpretation is consistent with the
per-feature ablation (Appendix Figure~\ref{fig:perfeat}).




\section{Conclusion}
\label{sec:conclusion}

We asked whether the added complexity of a temporal-graph anomaly detector
model is justified across five public datasets and one generated dataset. We
test whether a nonlinear model of all 14 features improves over selecting one
feature, how much accuracy \midas{} and \slade{} gain over a reference with no
parameter fitting, and how much compute they spend for that gain.
The reference exceeds \isoforest{} on all six datasets, matches or exceeds
\slade{} on three, and is within 13.7 percentage points of \slade{} on all
six. \slade{} requires 23 to 133 times the reference wall time, with a mean
of $72\times$ across all six datasets. Across the three \slade{} wins, the
means are $91\times$ and $+10.66$\,pp AUC. On the synthetic datasets, every
method remains near chance, while pre-event structural scores recover planted
patterns at AUC up to $0.955$.

Complexity earns its cost on the small Bitcoin graphs, where \slade{} recovers
signal absent from the selected feature. Across all six datasets, its mean
wall-clock cost is $72\times$ that of \simplebase{}. The benefit of complexity
is therefore dataset-dependent. We recommend that every claimed gain be
reported against a strong one-feature reference together with its compute cost.

\section*{Acknowledgments}
We thank Sudeep Agarwal, Greg Dunham, Mukul Sabharwal, and Euan Grant
(Microsoft) for leadership support. We further acknowledge the MAI / Bing
Fundamentals, Bing Defence, and Bot Detection teams at Microsoft for the
production-systems context that motivated this audit. Compute provided by
Microsoft.

\paragraph{Disclaimer.}
Some of the information in this document relates to pre-released content which
may be subsequently modified. Microsoft makes no warranties, express or
implied, with respect to the information provided here. This document is
provided ``as-is''. Information and views expressed in this document, including
URL and other Internet Web site references, may change without notice. Some
examples depicted herein are provided for illustration only and are fictitious.
No real association or connection is intended or should be inferred. This
document does not provide you with any legal rights to any intellectual
property in any Microsoft product.
\textcopyright\ 2026 Microsoft. All rights reserved.

\bibliographystyle{unsrtnat}
\bibliography{refs}

\appendix
\setcounter{figure}{0}
\renewcommand{\thefigure}{A\arabic{figure}}
\setcounter{table}{0}
\renewcommand{\thetable}{A\arabic{table}}

\section{Recommendations for benchmark authors}
\label{sec:recommendations}

Our audit suggests five additions to future temporal-graph anomaly-detection
evaluations.

\begin{enumerate}
\item \textbf{Report a chronologically selected one-feature baseline.} A
single feature selected from a fixed pool of counts, recencies, novelty
indicators, and count-derived transforms computed from prior interactions
(Section~\ref{sec:methods:selection}) is a compact reference. It matched or
beat \slade{} on three of our six datasets. We recommend reporting it
alongside learned models.
\item \textbf{Report paired significance in addition to seed CIs.} The
paired DeLong test~\cite{delong1988comparing}, or a paired permutation test
when score direction is part of the reported statistic, turns
"method A scores higher" into "method A is significantly better on
this dataset" (Section~\ref{sec:methods:delong}).
\item \textbf{Publish wall-clock cost alongside AUC.} A method that
wins by $+10.66$\,pp at $91\times$ the compute is a different
proposition from one that wins at parity. Per-percentage-point cost
(Figure~\ref{fig:cost-benefit}) makes the trade-off explicit.
\item \textbf{Include a nonlinear full-feature control.} This distinguishes
gains from combining engineered features from gains due to a temporal-graph
architecture.
\item \textbf{Validate architectural claims on controlled synthetic
probes with generator-informed pre-event structural scores.} Our
Synth-Triangle probe with a common-neighbor score
(Section~\ref{sec:methods:synth}) shows that "our model
exploits $k$-hop structure" is a testable claim, not an assumption:
the planted structure was recoverable from prior edges (AUC $=0.666$)
yet was not recovered by \slade{}.
\end{enumerate}

\section{Destination-activity Gini}
\label{app:regime}

For each node $v$ that appears at least once as a destination, let
\[
c_v = \sum_{(u,w,t)\in E}\mathbf{1}[w=v]
\]
be its number of destination-side interaction events. Repeated interactions
are counted separately, and nodes that never appear as destinations are
excluded. For the $n$ observed destination nodes, we define
\[
G_{\mathrm{dst}}
=
\frac{
\sum_{i=1}^{n}\sum_{j=1}^{n}|c_i-c_j|
}{
2n\sum_{i=1}^{n}c_i
}.
\]
A value of $0$ indicates equal destination activity, while larger values
indicate greater concentration.

\section{Evaluation protocol}
\label{app:eval}

\textbf{Split.}
A chronological 85/15 train/test split is applied to each dataset. The
simple-feature pipeline receives no parameter tuning beyond feature selection
on the chronological training slice (Section~\ref{sec:methods:selection}).

\textbf{Metric.}
Test-slice effective ROC-AUC is reported for all four methods in
Table~\ref{tab:master}. ROC-AUC is the standard metric in temporal-graph
anomaly detection;
Appendix~\ref{app:auprc} gives a complementary precision-recall analysis.
Effective AUC,
$\max(\text{AUC}_{\text{test}},1-\text{AUC}_{\text{test}})$, measures
orientation-invariant ranking separability rather than a prospectively
oriented deployment score.

\begin{figure}[t]
\centering
\includegraphics[width=\linewidth]{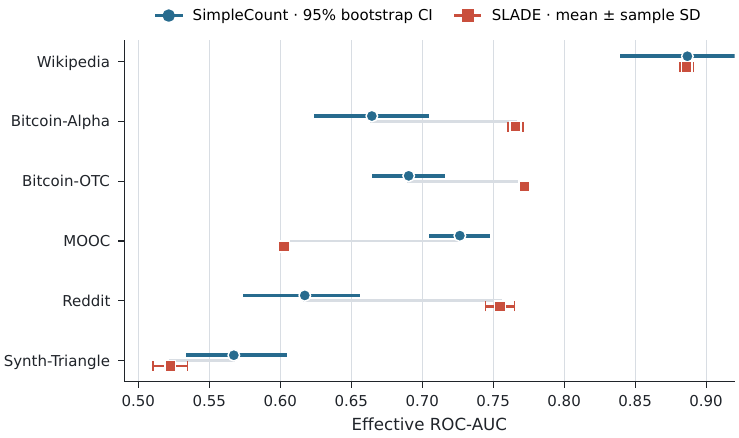}
\caption{Uncertainty and seed variation. Circles show \simplebase{}
with 95\,\% bootstrap CIs; squares show \slade{}'s five-seed
mean $\pm$ sample standard deviation. Gray connectors emphasize the
per-dataset gap. \slade{} sample standard deviation ranges from
$0.002$ to $0.012$.}
\label{fig:multiseed}
\end{figure}

\textbf{Paired testing.}
DeLong correlated-ROC tests \cite{delong1988comparing} use per-edge
scores from both methods on the same test slice, as described in
Section~\ref{sec:methods:delong}.
For Reddit, we use a paired, orientation-invariant permutation test. We
independently convert both score vectors to
normalized midranks, randomly swap the paired method labels, and
recompute the effective-AUC gap for each of $9{,}999$ permutations. The
\simplebase{} intervals in Figure~\ref{fig:multiseed} use 500 bootstrap
iterations, with per-iteration subsampling to $\le 100\,000$ edges for
computational tractability on the largest datasets (Reddit: 672\,447 edges).

\textbf{Multiple-testing correction.}
We apply Holm--Bonferroni correction~\citep{holm1979simple} with a family-wise
$\alpha = 0.05$ across the six \slade{}-vs-\simplebase{}
comparisons (one per dataset).

\section{Reddit paired permutation test}
\label{app:reddit-permutation}

We compare \simplebase{} and \slade{} with a paired Monte Carlo permutation
test that recomputes the orientation-invariant effective-AUC gap. Scores are
converted to normalized within-method midranks. In each of
$B=9{,}999$ replicates, the two ranked scores are independently exchanged
within each test edge with probability $1/2$, using random seed 20260730. With
the standard add-one correction, the two-sided $p$-value is

\[
p = \frac{E+1}{B+1},
\]

where $E$ is the number of sampled gaps at least as large in absolute value as
the observed gap. For Reddit, $E=26$, giving
$p=(26+1)/(9{,}999+1)=0.0027$.

\section{GraphMixer comparison}
\label{app:graphmixer}

We evaluate GraphMixer using the DyGLib dynamic-node-classification pipeline:
temporal-link pretraining followed by a node classifier trained on anomaly
labels. GraphMixer trails \simplebase{} by $2.3$ percentage points on Wikipedia
and leads by $2.1$ points on Reddit (Table~\ref{tab:graphmixer}).

\begin{table}[h]
\caption{GraphMixer comparison. Values are effective ROC-AUC; GraphMixer is
mean $\pm$ sample standard deviation over five seeds, and gap is GraphMixer
minus \simplebase{}.}
\label{tab:graphmixer}
\centering
\small
\begin{tabular}{lrrr}
\toprule
\textbf{Dataset} & \textbf{\simplebase{}} & \textbf{GraphMixer}
& \textbf{Gap} \\
\midrule
Wikipedia & $0.887$ & $0.864\pm0.014$ & $-2.3$\,pp \\
Reddit    & $0.618$ & $0.639\pm0.012$ & $+2.1$\,pp \\
\bottomrule
\end{tabular}
\end{table}

\section{Precision-recall robustness}
\label{app:auprc}
ROC-AUC can be optimistic under extreme class imbalance, so we report
average precision (AUPRC) on the same chronological test slices
(Table~\ref{tab:auprc}). \midas{}, \isoforest{}, and \simplebase{} use the
same deterministic scores as the ROC-AUC comparison; \slade{} values
are five-seed means and the chance level equals the test-slice base rate.
The three-of-six count from Table~\ref{tab:master} is unchanged under
AUPRC, but membership changes: \slade{} leads Wikipedia,
Bitcoin-Alpha, and Reddit, while \simplebase{} leads Bitcoin-OTC, MOOC,
and Synth-Triangle. The two highest test base rates are Bitcoin-OTC
($0.088$) and Bitcoin-Alpha ($0.064$). Bitcoin-OTC, a \slade{} win on
ROC-AUC, reverses in \simplebase{}'s favour under AUPRC. Wikipedia and
Reddit sit near the base-rate floor, where every method's AUPRC is below
$0.02$ and absolute differences are small. \simplebase{} exceeds
\isoforest{} on five datasets; on Reddit, \isoforest{} scores $0.00124$
versus $0.00118$ for \simplebase{}, an absolute difference below $0.0001$.

\begin{table}[h]
\caption{Average precision (AUPRC) on the chronological test slice.
Bold indicates the per-dataset best across the four methods; \slade{} is the
five-seed mean and chance equals the test-slice base rate. The
three-of-six pattern of Table~\ref{tab:master} is preserved.}
\label{tab:auprc}
\centering
\small
\begin{tabular}{lrrrrr}
\toprule
\textbf{Dataset} & \textbf{Chance} & \textbf{\midas{}}
& \textbf{\isoforest{}} & \textbf{\simplebase{}} & \textbf{\slade{}} \\
\midrule
Wikipedia      & 0.002 & 0.002 & 0.004 & 0.015 & \textbf{0.016} \\
Bitcoin-Alpha  & 0.064 & 0.102 & 0.097 & 0.111 & \textbf{0.148} \\
Bitcoin-OTC    & 0.088 & 0.146 & 0.128 & \textbf{0.289} & 0.205 \\
MOOC           & 0.009 & 0.019 & 0.013 & \textbf{0.023} & 0.016 \\
Reddit         & 0.001 & 0.001 & 0.001 & 0.001 & \textbf{0.003} \\
Synth-Triangle & 0.021 & 0.021 & 0.021 & \textbf{0.026} & 0.023 \\
\bottomrule
\end{tabular}
\end{table}

\section{Supplementary figures}
The following figures support the analyses in the main text.

\makeatletter
\setlength{\@fptop}{0pt}
\makeatother

\begin{figure}[t]
\centering
\includegraphics[width=\linewidth]{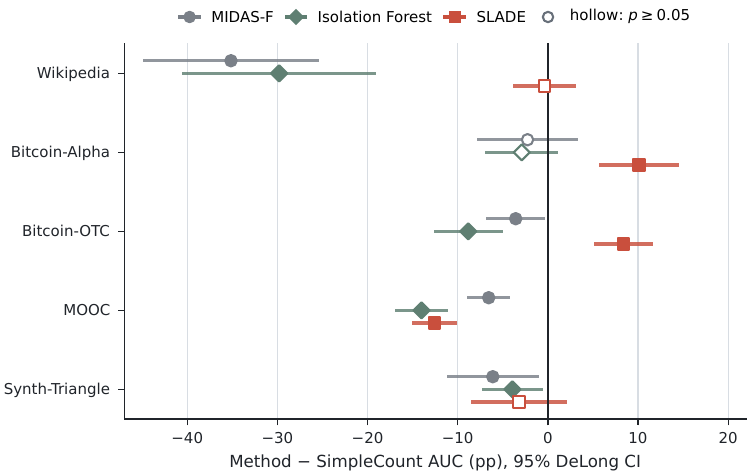}
\caption{Paired DeLong forest for \midas{}, \isoforest{}, and \slade{}
on five datasets. Reddit uses the paired permutation test and is omitted.
Intervals show effect size (method AUC $-$ \simplebase{} AUC) with 95\,\%
CIs; hollow markers denote unadjusted $p\geq0.05$.}
\label{fig:delong}
\end{figure}

\begin{figure}[t]
\centering
\includegraphics[width=\linewidth]{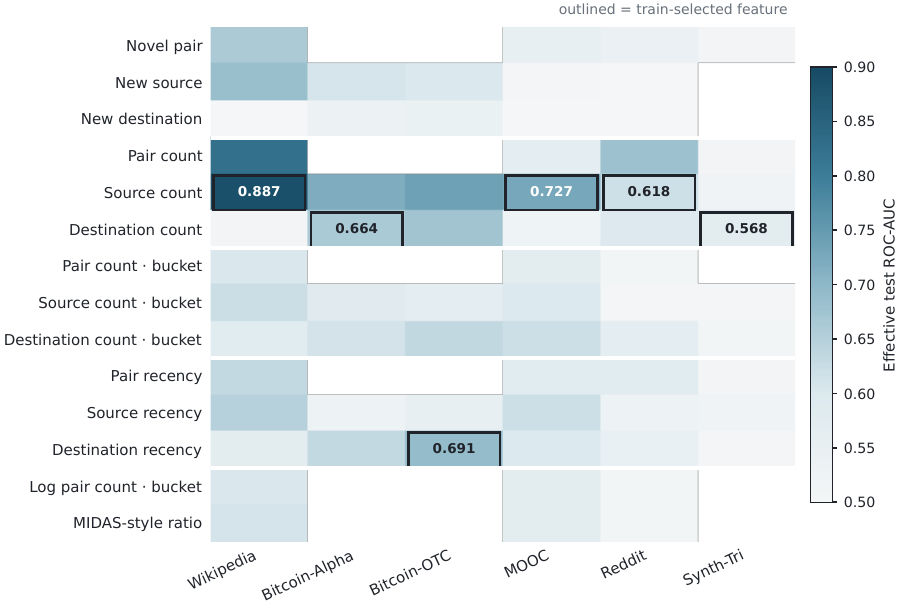}
\caption{Effective test AUC for each chronological candidate feature. Outlined
cells identify the feature selected on the chronological training
slice; their values are printed. Blank cells indicate that the feature
is constant on that test slice, so ROC-AUC is undefined. The selected
feature varies by dataset and is a per-source or per-destination count
or recency feature.}
\label{fig:perfeat}
\end{figure}

\section{Generative AI Usage Statement}
Generative AI tools were used to assist with language editing, code
development, and manuscript organization. 

\end{document}